\documentclass{article}
\usepackage{spconf,amsmath,graphicx,amssymb,citesort}
\pdfoutput=1

\usepackage{textcomp}
\usepackage{fancyhdr}
\usepackage[algoruled,linesnumbered]{algorithm2e}
\PassOptionsToPackage{bookmarks=false}{hyperref}
\makeatletter
\usepackage{epstopdf}
\ninept
\twoauthors
{Jacob Chakareski, Immanuel Manohar\sthanks{JC and IM are supported by NSF grant CCF-1528030.}}
{%Electrical and Computer Engineering\\
University of Alabama, Tuscaloosa, AL 35487.}
{Shantanu Rane}
{Xerox PARC, Palo Alto, CA 94304.}

\makeatother

\begin{document}

\title{Matrix Factorization-Based Clustering of Image Features\\ for Bandwidth-Constrained Information Retrieval}
\maketitle
\begin{abstract}
% !TEX root = ../MAIN.tex
We consider the problem of accurately and efficiently querying a remote server to retrieve information about images captured by a mobile device. In addition to reduced transmission overhead and computational complexity, the retrieval protocol should be robust to variations in the image acquisition process, such as translation, rotation, scaling, and sensor-related differences. We propose to extract scale-invariant image features and then perform clustering to reduce the number of features needed for image matching. Principal Component Analysis (PCA) and Non-negative Matrix Factorization (NMF) are investigated as candidate clustering approaches. The image matching complexity at the database server is quadratic in the (small) number of clusters, not in the (very large) number of image features.  We employ an image-dependent information content metric to approximate the model order, i.e., the number of clusters, needed for accurate matching, which is preferable to setting the model order using trial and error.  We show how to combine the hypotheses provided by PCA and NMF factor loadings, thereby obtaining more accurate retrieval than using either approach alone. In experiments on a database of urban images, we obtain a top-1 retrieval accuracy of 89\% and a top-3 accuracy of 92.5\%.  
\end{abstract}

% !TEX root = ../MAIN.tex
\section{Introduction}
\label{sec:intro}

\begin{figure}[t]
\centering
\includegraphics[width=2.5in]{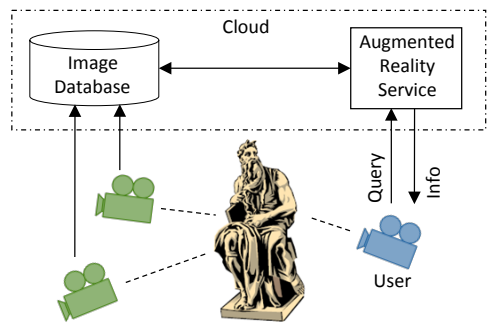}
\caption{\label{fig:Illustration} Image-based information retrieval for augmented reality.}
\end{figure}

Image-based information retrieval is becoming an important component of several technologies, including car navigation, video surveillance,
mobile augmented reality and many more. A camera, usually installed on a mobile device, captures an image of the scene of
interest and sends the image, or features extracted from the image to a remote server. The server compares the received features
against its own database of images. Depending upon the application, the server sends back to the mobile device: matching images of the scene,
or scene metadata, or control instructions to be executed at the mobile device. While accurate information retrieval is
important, designing such systems involves negotiating an appropriate tradeoff amongst several other variables.
The transmission overhead, computational complexity at the mobile device and at the database server, and robustness to
variations in the image acquisition process, are some of the important considerations that need to be balanced.
Figure \ref{fig:Illustration} shows an augmented reality use-case in which the query image might not accurately match any of the images
in the database owing to scaling variations, shifts, rotations and occlusions in the scene.

% talk about sift, surf, brisk, freak with references.
To achieve robustness to variations in the image acquisition process, it is customary to use a feature space that is invariant to these variations. The
most popular feature set is the Scale Invariant Feature Transform (SIFT)~\cite{lowe2004distinctive}, which produces descriptors that are
robust to rotation and uniform scaling, and partially invariant to affine distortion and illumination effects. While we use SIFT in our study, the underlying concepts should extend to other features
based on ``keypoints'', i.e., salient locations in the image. These include SURF~\cite{bay2006surf}, HoG~\cite{dalal2005histograms}, CHoG~\cite{ChoG},
BRISK~\cite{leutenegger2011brisk}, FREAK~\cite{alahi2012freak} and others. To find an image in the server's database that matches the image captured by the mobile device,
it is necessary to compare features from the query image against those from the database images. \

One way to reduce the complexity of the matching
process is to reduce the dimensionality of the feature vectors without compromising their matching ability. This is the approach followed in
PCA-SIFT which obtains SIFT key points and then employs principal component analysis to reduce the dimensionality of the image patch
around the key point~\cite{ke2004pca}. Another class of methods to reduce the dimensionality of the image features is inspired by Locality Sensitive
Hashing~\cite{indyk98stoc}. These methods involve computing one-bit random projections of the image features, and then matching the images in the subspace of
random projections~\cite{RPDSCPCA}. These methods were later generalized in~\cite{li2012quantized}, in which a quantized version of the Johnson-Lindenstrauss transform,
was shown to improve matching performance by trading off the quantization step-size against the dimensionality of the projected features.

%Many other descriptors soon followed namely, SURF, HoG, etc.., \cite{bay2006surf,dalal2005histograms}.  These image features were high dimensional, example, SIFT descriptors are 128 dimensional vectors and has a huge number of descriptors, example, for Zurich Buildings database, \cite{shao2003zubud}, the algorithm proposed in \cite{lowe2004distinctive}, identifies on an average, 2253 keypoints for one image. The compression of SIFT features directly using PCA, ICA and cHoG was studied in \cite{chandrasekhar2010survey, gonzalez2012dimensionality}.

% talk about  dimensionality reduction and drawbacks of reducing dimensionality but not that number of key points.

In this paper, we consider a different approach to dimensionality reduction than the ones discussed above\footnote{We are interested in
techniques that do not require training of a ``good'' feature set. Training-based
methods~\cite{SmallC,SpectralH,PCAADC,LDAHash} are very accurate, but they can also become cumbersome when the database keeps growing.
When new landmarks, products, etc. are added, feature statistics are altered
and the trained feature set must be updated.}. Concretely, rather than reducing the dimension of each individual feature vector, we seek to reduce the total number of feature vectors per image. This is driven by the observation that, for good retrieval performance, a large number of keypoint-based features have
to be extracted and used for matching. For the image sizes encountered today, this number can easily range from a few hundred to a few thousand
descriptors. We compare two matrix factorization-based approaches to reduce the number of descriptors, viz., Principal Component Analysis (PCA) 
and a sparse Non-negative Matrix Factorization (NMF) approach developed for video querying~\cite{mansour2014video}. Our choice is motivated by
the fact that PCA and NMF factor loadings are closely related to cluster centroids that would be obtained if $k$-means clustering was applied to the image
features~\cite{ding2004k,ding2006orthogonal}. In doing so, we encounter a new problem: How many PCA or NMF factor loadings -- equivalently, how many clusters -- are enough for good matching
performance? The answer to this question is image-dependent, and must be known to the mobile device. We employ an estimate of information content,
previously used in econometric studies~\cite{bai2002determining} to approximate the number of PCA and NMF factor loadings that will be
used for matching. This estimate turns out to be significantly smaller than the number of keypoint-based features extracted per image, and
incurs little or no penalty in matching performance.

An additional challenge presents itself at the database server: How does one match the basis vectors received from the mobile
device against the set of basis vectors extracted from each database images. A natural solution is to correlate individual PCA or NMF factor loadings
of the query image with those of the database images. While this approach works well for video querying~\cite{mansour2014video}, it might not
provide good matching performance when the objects in the server's database have been photographed from vastly different viewpoints.
We investigate a second approach which is based on the angle between the subspaces
spanned by the PCA or NMF factor loadings of the query image and those of the database images. We report our findings for both matching criteria,
evaluated on a database of urban images~\cite{shao2003zubud}. Furthermore, we show how to improve the accuracy of image-based information retrieval by combining the hypotheses
obtained from the PCA and NMF bases.

The remainder of this paper is organized as follows: In Section \ref{sec:SystemArchitecture}, we fix notation and provide a brief description of the main feature clustering approaches investigated in this paper. In Section \ref{sec:method}, the proposed image retrieval algorithm is described. The computational complexity of our approach is investigated in Section \ref{sec:complexity}. Our experimental results obtained using a database of urban images, are detailed in Section \ref{sec:res}.

\section{Background and Notation}
\label{sec:SystemArchitecture}
% !TEX root = ../MAIN.tex

We now briefly describe the building blocks of our image retrieval scheme and also set the notation that will be used throughout the
paper. The main building blocks include the feature extraction and feature clustering schemes based on PCA and NMF.

\subsection{Feature extraction}
\label{sub:Descriptor_Construction}

As described earlier, the mobile device extracts keypoint-based features from the captured image. Similarly, the database server
extracts keypoint-based features from each of its images. Let the total number of keypoints in a given image be $N$ and let $\mathbf{d}_{i}\in\mathbb{R}^{T}$ be
the descriptor, or feature vector, corresponding to the $i^{th}$ keypoint. The descriptors can then be stacked to form a $T \times N$ matrix
$\mathbf{M} \triangleq\left[\mathbf{d}_{1},\,\mathbf{d}_{2},\,\ldots \mathbf{d}_{N}\right]$.

For concreteness, we focus on SIFT features henceforth, though other feature
spaces are also usable within our framework. We frame the problem of computing a compact descriptor for an image as
that of finding a low-dimensional representation of the matrix $\mathbf{M}$. Below, we discuss two ways for constructing such a representation.

\subsection{PCA-based feature clustering}
\label{sub:PCA}

Using PCA, the matrix $\mathbf{M}$, can be written as $\mathbf{M}=\mathbf{HF}+\mathbf{E}$, where
$\mathbf{H}\in\mathbb{R}^{T\times k}$ is called the factor loading matrix, $\mathbf{F}\in\mathbb{R}^{k\times N}$ is a matrix whose columns
serve as PCA factors and $\mathbf{E}\in\mathbb{R}^{T\times N}$ is a noise matrix with covariance $\sigma^{2}\mathbf{I}$~\cite{bai2002determining}.
The pair $\left(\mathbf{H},\,\mathbf{F}\right)$ is not unique and can be replaced by the pair $\left(\mathbf{HQ},\,\mathbf{Q^{-1}F}\right)$
for any non-singular matrix $\mathbf{Q}$ but the space spanned, given by $\mathcal{R}\left\{ \mathbf{H}\right\} $
remains constant. To determine $\mathbf{H}$ and $\mathbf{F}$, we compute the Singular Value Decomposition
of $\mathbf{M}$, given by $\mathbf{U} \mathbf{\Sigma}\mathbf{V}^{\top},$ where $\mathbf{U}$ and $\mathbf{V}$ are orthogonal matrices
and $\mathbf{\Sigma}$ is a $T \times N$ matrix containing singular values
%given by \[
%\Sigma=\left[\begin{array}{ccccc}
%\sigma_{1} &  & \ldots\\
% & \sigma_{2} &  &  & 0\\
% &  & \ddots\\
% &  &  & \sigma_{T}
%\end{array}\right]_{T\times N}
%\]
$\sigma_{1} \geq \sigma_{2}\geq \ldots \geq \sigma_{T}$ on its diagonal and zeros everywhere else.
If the model order, or the dimension of the space spanned by $\mathbf{H}$ is known to be $k$, then $\mathbf{H}$ is obtained
simply by taking the first $k$ columns of $\mathbf{U}$, which are the singular vectors corresponding to the $k$ highest
singular values. In practice $k$ is unknown and we will estimate it as described in the next section.

\subsection{NMF-based feature clustering}
\label{sub:NMF}

Using NMF,  the matrix $\mathbf{M}$, can be written as $\mathbf{M}=\mathbf{LR}$, where $\mathbf{L}\in\mathbb{R}^{T\times k}$
here corresponds to the factor loading matrix and $\mathbf{R}\in\mathbb{R}^{k\times N}$ is the matrix whose columns serve as
NMF factors. We employ the technique used in \cite{mansour2014video}, where NMF was used to cluster video descriptors.
SIFT features were extracted from consecutive frames in the video sequence and stacked into a matrix similar to $\mathbf{M}$,
containing all non-negative elements. By design, both $\mathbf{L}$ and $\mathbf{R}$ are constrained to have all non-negative elements.
There exist many flavors of NMF algorithms~\cite{gillis2014and,hoyer2004non}. We adopt the sparse NMF algorithm
proposed in \cite{mansour2014video} which constrains each column of $\mathbf{R}$ to belong to the set of standard basis vectors:
\[
E_{r}=\left\{ \mathbf{e}_{j}\in\mathbb{R}^{k}:\,e(j)=1,\mbox{and 0 otherwise},\,j\in\left\{ 1,\,\ldots\,k\right\} \right\} .
\]
 The optimization problem set up to determine $\mathbf{L}$ and $\mathbf{R}$ is given by:
\begin{align*}
(\widehat{\mathbf{L}},\,\widehat{\mathbf{R}}) & = \min_{\begin{array}{c}
\mathbf{L}\in\mathbb{R}^{T\times k}\\
\mathbf{R}\in\mathbb{R}^{k\times N}
\end{array}}\frac{1}{2}\left\Vert \mathbf{M-LR}\right\Vert _{F}^{2},\\
\text{subject\,to:} \ & \begin{cases}
\left\Vert \mathbf{L}_{i}\right\Vert _{2}=1, & \forall i\in\left\{ 1,\,\ldots,\,k\right\} \\
\left\Vert \mathbf{R}_{j}\right\Vert _{0}=1, & \forall j\in\left\{ 1,\,\ldots,\,N\right\}
\end{cases}.
\end{align*}
In~\cite{mansour2014video}, the model order $k$ was decided \emph{a priori}. While we use the same NMF algorithm, the situation
differs in two aspects. Firstly, the procedure is applied to features extracted from a single image, rather than to features extracted from
multiple image frames. Secondly, the model order is not determined ahead of time and is allowed to vary depending on the image
content. For estimating the model order, we rely on an estimate of information content that is computed from the PCA
decomposition of $\mathbf{M}$. Determination of the model order is discussed in the next section.

\section{Proposed image-based retrieval scheme}
\label{sec:method}
% !TEX root = ../MAIN.tex

\begin{figure}
\begin{centering}
\includegraphics[width=8cm]{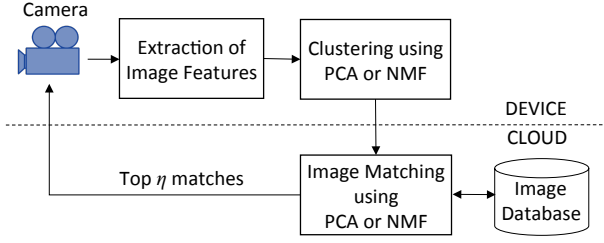}
\par\end{centering}
\caption{\label{fig:system_description} The database server identifies the object (or relevant information about the object) photographed by the client
device by matching a few compact image descriptors derived from a large number of scale-invariant image features.}
\end{figure}

A block diagram of our image-based information retrieval scheme is shown in Figure \ref{fig:system_description}. First, a client device captures a query image and extracts keypoint-based image features from it. As explained earlier, the client then compresses the feature space using PCA or sparse NMF. The PCA or NMF factor loadings, representing ``compressed'' features, are sent to the server where they are matched against the factor loadings extracted from images in the server's database.  Information about the top $\eta$ matching objects is returned to the client. Below, we first describe the
database preparation process that is performed at the server. We also explain how the model order $k$, i.e., the number of relevant clusters,
is chosen for each image at both the client and the server. Finally, we describe how the server determines the object in its database that most
closely matches the query image.

% Subsection \ref{sub:Retrieval_framework} describes the retrieval framework used here in detail, followed by Subsection \ref{sub:model_order} which describes the optimal dimension of the features when using PCA and NMF, Subsection \ref{sub:Database_Retrieval} describes the techniques used to match the compressed image features with the database to generate the top $\eta$ matches.

\subsection{Client-side and server-side processing}
\label{sub:Retrieval_framework}
Let $I_{q}$ be the query image captured by the client. As explained in Section~\ref{sec:SystemArchitecture}, image features extracted from the query image are stacked together to obtain a matrix $\mathbf{M}_q$,
which is compressed to a $T \times k$ matrix of PCA bases  $\mathbf{H}_q$, or to a $T \times k$ matrix of NMF bases  $\mathbf{L}_q$. Depending upon
which matrix factorization technique is used $\mathbf{H}_q$, or $\mathbf{L}_q$, or both are sent to the server.

Let the server's database consist of $K$ images, given by the set $\left\{ I_1,\,I_2,\ldots,\,I_K\right\}$. Let the corresponding
PCA representations be $\mathbf{\mathcal{H}}=\left\{ \mathbf{H}_1,\,\mathbf{H}_2,\,\ldots,\,\mathbf{H}_K\right\}$,
and the NMF representations be $\mathbf{\mathcal{L}}=\left\{ \mathbf{L}_1,\,\mathbf{L}_2,\,\ldots\,\mathbf{L}_K\right\}$. To compute
the best matching images, according to the PCA representations, we compare $\mathbf{H_q}$ against
$\mathbf{\mathcal{H}}$. We represent the top $\eta$ matches obtained using the PCA representation as the set
$\mathcal{V}_{PCA}=\left\{ h_1, h_2, ..., h_{\eta}\right\}$, where $h_1$ represents the best match and $h_{\eta}$ the
worst. If the server has multiple images per object, then the list should contain the $\eta$ best matching \emph{objects}, rather than
the $\eta$ best matching \emph{images}. To achieve this, we examine the list $\mathcal{V}_{PCA}$, and if it contains more than one
image of a given object, we retain only the best matching image and remove the others. 

Similarly, for the NMF representations, we compare $\mathbf{L_q}$ against
$\mathbf{\mathcal{L}}$, and represent the top $\eta$ matches as the set
$\mathcal{V}_{NMF}=\left\{ \ell_1, \ell_2, ..., \ell_{\eta}\right\}$. The two hypotheses, $\mathcal{V}_{PCA}$ and $\mathcal{V}_{NMF}$ can be
combined to get a more refined set of top-$\eta$ matches. Later, we provide a heuristic algorithm for this purpose.

%\begin{equation}
%\mathbf{\mathcal{H}}=\left\{ \mathbf{H_{1}},\,\mathbf{H_{2}},\,\ldots,\,\mathbf{H_{K}}\right\} \label{eq:PCA_server}
%\end{equation}  correspond to the set of PCA representations and \begin{equation}
%\mathbf{\mathcal{L}}=\left\{ \mathbf{L_{1}},\,\mathbf{L_{2}},\,\ldots\,\mathbf{L_{K}}\right\}, \label{eq:NMF_server}
%\end{equation}  corresponds to the NMF representations. The query's $\mathbf{H_q}$ is matched $\mathbf{\mathcal{H}}$ for the database images to retrieve the first $\eta$ matches in order to form the PCA hypothesis, $\mathcal{V}_{PCA}=\left\{ m_{H1},\,m_{H2},\,\ldots,\,m_{H\eta}\right\} $ where $m_{H1}$ corresponds to the first best match and $m_{H\eta}$ corresponds to the $\eta^{th}$ best match. For these $\eta$ images, $L_i$ such that $i\in \mathcal {V}_{PCA}$  is matched with query's $L_q$ to form the order of matches which represents the NMF Hypothesis, $\mathcal{V}_{NMF}=\left\{ m_{L1},\,m_{L2},\,\ldots,\,m_{L\eta}\right\}$. The measures used to compare the matrices to form these hypothesis is discussed in \ref{sub:Comparison_metrics}.
%

\subsection{Determining the model order}
\label{sub:model_order}
An important consideration in our scheme is the model order, or the number of PCA/NMF
factor loadings, or the number of feature clusters. This is the value of $k$ (Section~\ref{sec:SystemArchitecture}) needed for accurate matching. If $k$ is too
small, the matching accuracy will suffer. If it is too large, then the transmission and computational overhead of
the protocol can become prohibitive. A natural way to ascertain the model order is to examine the singular values of $\mathbf{M}_q$.
If the first $k$ singular values are large, while the remaining singular values are small, then it makes sense to truncate the model order to $k$.
This is indeed the case for photographs of many natural and urban scenes. This motivates a technique, originally used for
determining the information content in econometric data in \cite{bai2002determining}, which we describe briefly below.

%When using PCA to model the image descriptors as in \eqref{eq:factor_model} or when modelling NMF as in \eqref{eq:NMF_model}, it's of interest to know if there exists an optimal $k$ for each image. Note, if the model \eqref{eq:factor_model} holds true for the image descriptors, then, the knowledge of $k$ can be based on the eigenvalues of $\mathbf{M}$, \eqref{eq:SIFT_descriptors}, as the first $k$ eigenvalues of $\mathbf{M}$ remains bounded in $O(1)$ whereas the rest of the eigenvalues tends to zero as the number of image descriptors $N \rightarrow\infty$. It was noted by experimentation on the ZuBuD database of images when using the SIFT descriptors, \cite{lowe2004distinctive}, as the number of descriptors were increased, while a few highest singular values converged to a constant value greater than zero, remaining set of lower singular values converged towards zero. It was also noted that the noise term in \eqref{eq:factor_model} was independent of each other when modeling the SIFT descriptors in ZuBuD database. Under these conditions, for a finite $N$, the Information theoretic criteria
%proposed in \cite{bai2002determining} could be used to estimate $k$.
For an image containing $T$-dimensional features extracted from $N$ key points, the \emph{Information Content} captured in $k < \min(T,N)$
principal components is given by
\[
I(k)=\ln\left(V(k,\mathbf{F}^{(k)}_N)\right)+ k\left(\frac{T+N}{TN}\right)\ln\left(\frac{TN}{T+N}\right)
\]
\[
\text{where}\,\,\,V(k,\mathbf{F}^{(k)}_N)=\min_{\mathbf{\hat{H}}}\frac{1}{TN}\sum_{i=1}^{N}\left\Vert \mathbf{d}_i - \mathbf{\hat{H}}^{(k)}\mathbf{f}_i^{(k)}\right\Vert _{2}^{2}
\]

Here, $\mathbf{d}_i$ is the $i^{th}$ descriptor in the matrix $\mathbf{M}$ of stacked descriptors, $\mathbf{\hat{H}}^{(k)}$ is the factor loading matrix
obtained by assuming model order as $k$, $\mathbf{f}_i^{(k)}$ is the $i^{th}$ column of the factor matrix, $\mathbf{F}^{(k)}_N$. In the above relations, $\mathbf{\hat{H}}^{(k)}$
and $\mathbf{F}^{(k)}_N$ represent the factor loading matrix and the factor matrix obtained using PCA in subsection \ref{sub:PCA} under
the assumption that the model order was $k$. The estimate of correct model order is given by $k^*= \arg \min_{k} I(k)$.
%\begin{equation}
%k^*= \arg \min_{k} I(k).\label{eq:modelorder}
%\end{equation}

The above development allows us to estimate the order for PCA-based feature compression. To estimate the order for NMF-based feature compression,
we reason as follows: Suppose, we had performed $k$-means clustering of the image features, i.e., the columns of $\mathbf{M}$. 
We know that the subspace spanned by $k^*$ cluster centroids is the same as the subspace spanned by the first $k^* - 1$ columns of the PCA factor loading matrix $\mathbf{H}$~\cite{ding2004k}. Furthermore, it
has been remarked that NMF factor loadings, i.e., the columns of $\mathbf{L}$ are closely related to the $k$-means cluster centroids \cite{mansour2014video,ding2006orthogonal}. Because of this relationship between PCA factor loadings, $k$-means cluster centroids, and NMF factor loadings,
we choose the same model order $k^*$ for PCA-based and NMF-based feature compression.

\subsection{Comparing the query and database images}
\label{sub:Comparison_metrics}
For PCA representations, the server's task is to find a member of the set $\mathcal{H}$ that best matches the query factor loading
matrix $\mathbf{H}_q$. For NMF representations, it is to find a member of the set $\mathcal{L}$ that best matches the query factor loading
matrix $\mathbf{L}_q$. We now consider two matching criteria. Let $\mathbf{A}\in \mathbb{R}^{T \times k_{a}}$ and
$\mathbf{B}\in \mathbb{R}^{T\times k_{b}}$  be the two matrices to be compared. In our scheme, $\mathbf{A}$ might correspond to
either $\mathbf{L}_q$ or $\mathbf{H}_q$ and $\mathbf{B}$ might belong to either $\mathcal{L}$ or $\mathcal{H}$ respectively.

 The first metric we consider is the angle between subspaces spanned by $\mathbf{A}$ and $\mathbf{B}$~\cite{meyer2000matrix}.
 Let $\mathbf{P_{A}}$ and $\mathbf{P_{B}}$ be the projection matrices for $\mathbf{A}$ and $\mathbf{B}$ respectively, thus
$\mathbf{P_{A}}=\mathbf{A}\left(\mathbf{A}^{\top}\mathbf{A}\right)^{-1}\mathbf{A}^{\top}$. Then, the angle between the subspaces
spanned by $\mathbf{A}$ and $\mathbf{B}$ is given by,
\begin{equation}
\measuredangle \left(\mathbf{A},\,\mathbf{B}\right)=\cos^{-1}\left(\left\Vert \mathbf{P_{A}}\mathbf{P_{B}}\right\Vert _{2}\right).\label{eq:angle_NMF_subspaces}
\end{equation}
%For more details on angle between subspaces, refer section 5.15 in \cite{meyer2000matrix},

The second metric we consider is based on the maximum correlation between the columns of $\mathbf{A}$ and $\mathbf{B}$. To obtain this,
construct a matrix $\mathbf{S}\triangleq \mathbf{A}^{\top}\mathbf{B}\in\mathbb{R}^{k_{a}\times k_{b}}$. Then obtain 
the maximum value along each column of $\mathbf{S}$ and write the vector $\mathbf{s}_{max}$ such that $\mathbf{s}_{max}=[s_{1},\,\ldots,\,s_{k_{b}}],$
where, $s_{j}$ is the maximum value in the $j^{th}$ column of $\mathbf{S}$. The correlation score is then given by
\begin{equation}
\text{score}\left(\mathbf{A},\,\mathbf{B}\right)\triangleq\sum_{l=1}^{k_{b}}s_{l}.\label{eq:score_PCA_basis}
\end{equation}

\subsection{Combining the PCA and NMF hypotheses}
\label{sub:Combining_Classifiers}

We now describe how to improve the retrieval accuracy by combining the hypothesis of the matching image obtained via
the NMF and PCA-based feature clustering approaches. We choose the NMF retrieval result as the primary hypothesis and
the PCA retrieval result as the secondary hypothesis. Using the notation from Section~\ref{sub:Retrieval_framework}, we have
$\mathcal{V}_{pri} = \mathcal{V}_{NMF} = \left\{ \ell_1, \ell_2, ..., \ell_{\eta} \right\}$ and
$\mathcal{V}_{sec} = \mathcal{V}_{PCA} = \left\{ h_1, h_2, ..., h_{\eta} \right\}$.
To combine the primary and secondary hypothesis, an iterative
algorithm is designed. The order of any pair of retrieved objects in the
primary list is reversed if both the following conditions are satisfied: (a) The pair of objects appear
in the reverse order in the secondary list, and (b) if
the gap between the objects in the primary list
is exceeded by $0\leq\alpha\leq\eta$ places in the secondary list.
%Eg, $m_{p_{j}}$ and $m_{p_{j+l}}$ will be
%switched if $m_{p_{j+l}}=m_{s_{i}}$ and $m_{p_{j}}=m_{s_{i+l+\alpha}}$.
The parameter $\alpha$ serves as a relative weighting factor for the primary and secondary
hypotheses. Increasing $\alpha$ favors the primary hypothesis, and setting $\alpha=\eta$ completely ignores the
secondary hypothesis. A step-by-step procedure for combining the primary and secondary hypotheses
is provided in Algorithm \ref{alg:Combining-classifiers}.

\begin{algorithm}
\caption{\label{alg:Combining-classifiers}Combining PCA and NMF retrieval hypotheses.}
\KwData{$\alpha$ , $\mathcal{V}_{pri}$, $\mathcal{V}_{sec}$}
\KwResult{Resorted $\mathcal{V}_{pri}$}
 $i=1$, $j=1$, $permutations = 1$\;
\While{permutations = 1}{
$permutations = 0$\;
\While{$i< \eta/2 $}{
  \While{$j<= \eta -i$}{
   Find objects $h_a, h_b \in \mathcal{V}_{sec}$ that correspond to objects $\ell_i$
and $\ell_{i+j} \in \mathcal{V}_{pri}$ \;
   \If{$a+\alpha+j>b$}{
   switch $\ell_i$ and $\ell_{i+j}$\;
   $permutations = 1$
   }
   $j=j+1$
   }
   $i=i+1$
   }
   }
\end{algorithm}

\section{Complexity of Server-based Matching}
\label{sec:complexity}
% !TEX root = ../MAIN.tex
We assume that NMF-based and PCA-based
feature clustering has already been performed for all $K$ images in the
server's database. Suppose that there is an average of $N$ keypoint-based
features per image. To compute the angle between subspaces, we multiply two $T \times T$ projection matrices, 
and compute matrix norms. This incurs $O(T^3)$ complexity per image,
resulting in a total complexity of $O(KT^3)$. On the other hand, computing
the pairwise correlation between $T$-length columns of factor loading matrices with
model order $k_p$ and $k_q$ respectively, incurs $O(T k_p k_q)$ complexity.
Writing $k = \max(k_q, k_p)$, the total complexity for image matching based
on pairwise correlations is $O(TKk^2)$. In our experiments, described below, we
find that the angle between subspaces metric gives higher accuracy for NMF-based features,
while both similarity metrics work for PCA-based features.
To reduce the overall complexity, we use the correlation metric for PCA-based features
first to obtain $\eta$ matches, and then use the angle between subspaces metric with NMF-based features only on those
$\eta$ matching images. This brings the total complexity to $O(T K k^2 + \eta T^3)$. 
Effectively, the combined scheme (described above) amounts to performing PCA-based matching for a 
given list length $\eta$, reordering that list using NMF-based matching, and then~\emph{selectively
reversing} some of the reordering based on the parameter $\alpha$. 
Note that, as $\eta$ is very small (usually less than 20), we ignore the extra complexity of Algorithm~\ref{alg:Combining-classifiers}.

In comparison (See Table~\ref{table:Cplxity}), approaches that do not reduce the number of features incur
much higher matching complexity. For instance, to obtain the top-1 match, the
methods using direct matching of SIFT features, such as~\cite{li2012quantized}, incur
a complexity of $O(KN^2)$. To see why the proposed approach is significantly
less complex, recall that $N \gg k$ and $N \gg T$. E.g., in our experiments,
$T = 128$ for SIFT vectors, the average value of $k$ is 25, while $N = 2253$. 

\begin{table}
\begin{centering}
\begin{tabular}{|c|c|}
\hline
Scheme & Complexity\tabularnewline
\hline
\hline
Correlation between columns, given by (\ref{eq:score_PCA_basis}) & $O\left(Kk^{2}\right)$\tabularnewline
\hline
Angle between subspaces, given by (\ref{eq:angle_NMF_subspaces}) & $O\left(KT^{3}\right)$\tabularnewline
\hline
Proposed combined PCA + NMF scheme & $O\left(T K k^2 + \eta T^3\right)$\tabularnewline
\hline
Using (\ref{eq:score_PCA_basis}) without feature clustering & $O\left(KTN^{2}\right)$\tabularnewline
\hline
Methods described in \cite{chandrasekhar2010survey}, \cite{ke2004pca}, \cite{li2012quantized} & $O\left(KN^{2}\right)$\tabularnewline
\hline
\end{tabular}
\par\end{centering}

\caption{\label{table:Cplxity}Computational complexity of various schemes.}
\end{table}

\section{Experiments}
\label{sec:res}
\begin{figure}[t]
\begin{centering}
\includegraphics[width=8cm]{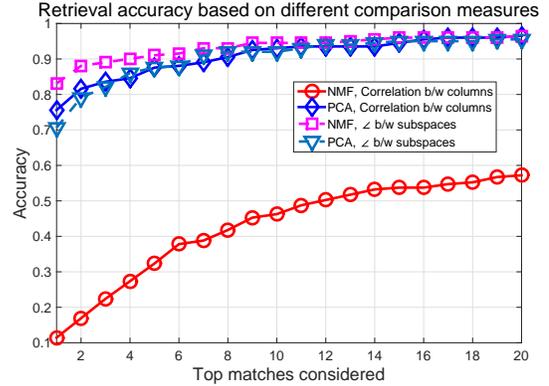}
\par\end{centering}
\caption{\label{fig:error_metrics}PCA-based retrieval has almost the same accuracy with both similarity metrics. For NMF-based retrieval,
the angle between subspaces metric is significantly more accurate.}
\end{figure}

% !TEX root = ../MAIN.tex
We use the Zurich Buildings Database (ZuBuD), \cite{shao2003zubud,shao2003hpat}, which consists of images of 201 buildings, each captured from 5 different viewpoints. The first image of each building was selected as the query image while the remaining 4 were
regarded as database images. The accuracy of retrieval, defined as the probability of a correct match, is used as a performance metric. 
SIFT descriptors -- 128 dimensions per descriptor -- were extracted from each image using the algorithm proposed in \cite{lowe2004distinctive} with 
the default recommended parameters. The factor loading matrix, denoted as $\mathbf{H}$, in the case of PCA, and $\mathbf{L}$, in the case of NMF, was constructed for each image based on its SIFT descriptors, as described in Section~\ref{sub:PCA} and Section~\ref{sub:NMF}, respectively. The $\mathbf{H}_q$ and $\mathbf{L}_q$ for a query image are then matched against the corresponding matrices in the database collections $\mathcal{H}$ and $\mathcal{L}$. The two similarity measures described in Section~\ref{sub:Comparison_metrics} are used to determine the correct match.
The resulting average matching accuracy for all 201 objects is shown in Figure~\ref{fig:error_metrics} with each similarity criterion.

It can be seen from Figure~\ref{fig:error_metrics} that for PCA-based matching, using $\mathbf{H}_q$ against $\mathcal{H}$, both metrics work well. 
The accuracy is slightly better when correlation amongst the columns is used as a similarity metric. However, this metric gives poor accuracy 
for NMF-based matching, using $\mathbf{L}_q$ against $\mathcal{L}$. This is in contrast to the superior performance observed 
in~\cite{mansour2014video}. We conjecture that this difference in performance is due to the difference in the type of data. 
In~\cite{mansour2014video}, the descriptors are constructed out of successive frames of a video sequence, while here, the descriptors
are constructed out of vastly differing viewpoints of the same scene. As shown in Figure~\ref{fig:error_metrics}, the NMF-based matching
gives much higher accuracy when the similarity metric used is the angle between the subspaces.
Measuring the angle between subspaces spanned by the NMF factor loadings of the query and database images, is akin to measuring the discrepancy between
the subspaces spanned by the centroids of the query features and the database features.
In all subsequent experiments, the comparison of $\mathbf{L}_q$ with $\mathcal{L}$ is carried out using the angle between subspaces and the comparison of $\mathbf{H}_q$ with $\mathcal{H}$ is carried out by evaluating the maximum correlation between the columns of the matrices.

\begin{figure}[t]
\begin{centering}
\includegraphics[width=8cm]{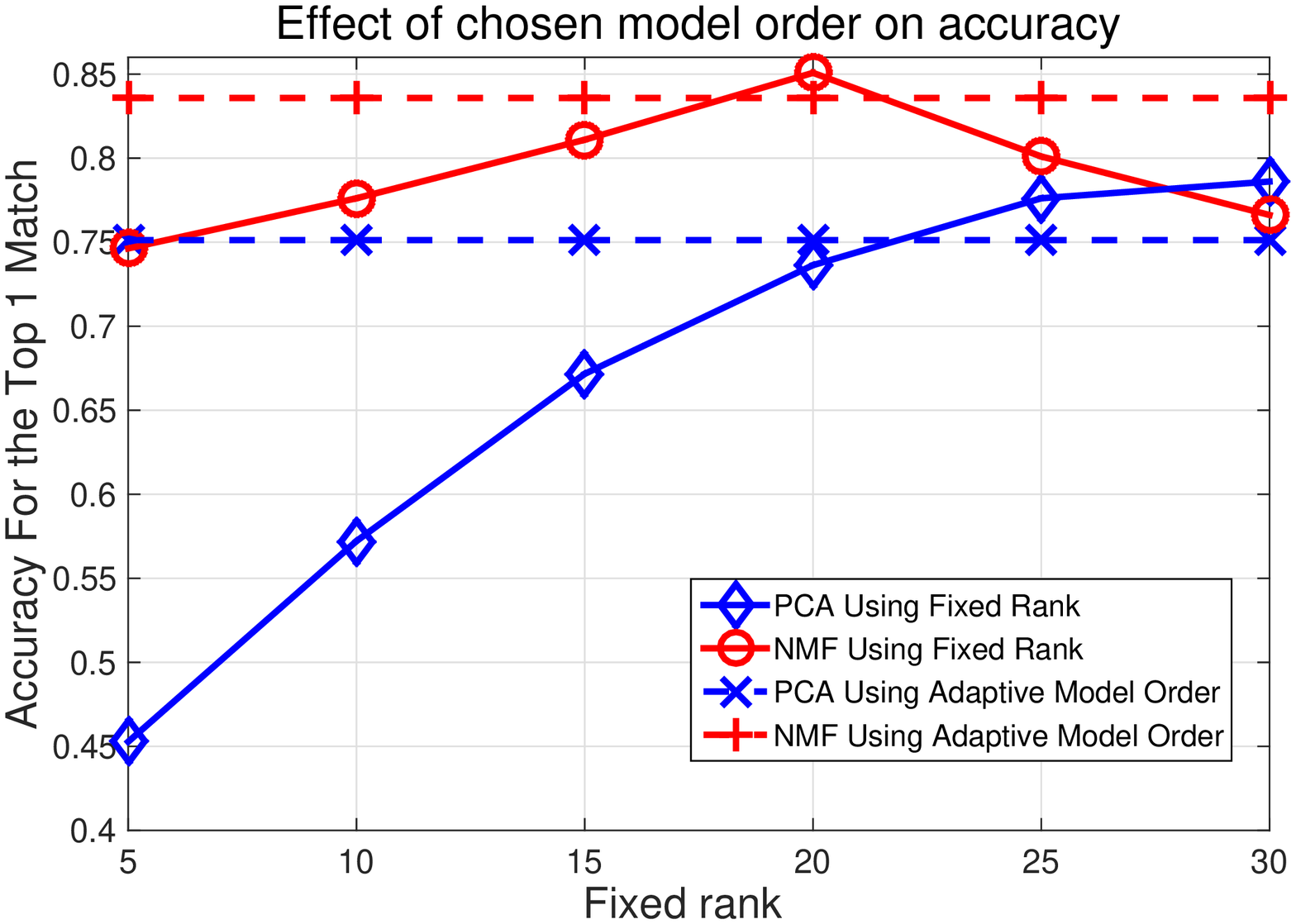}
\includegraphics[width=8cm]{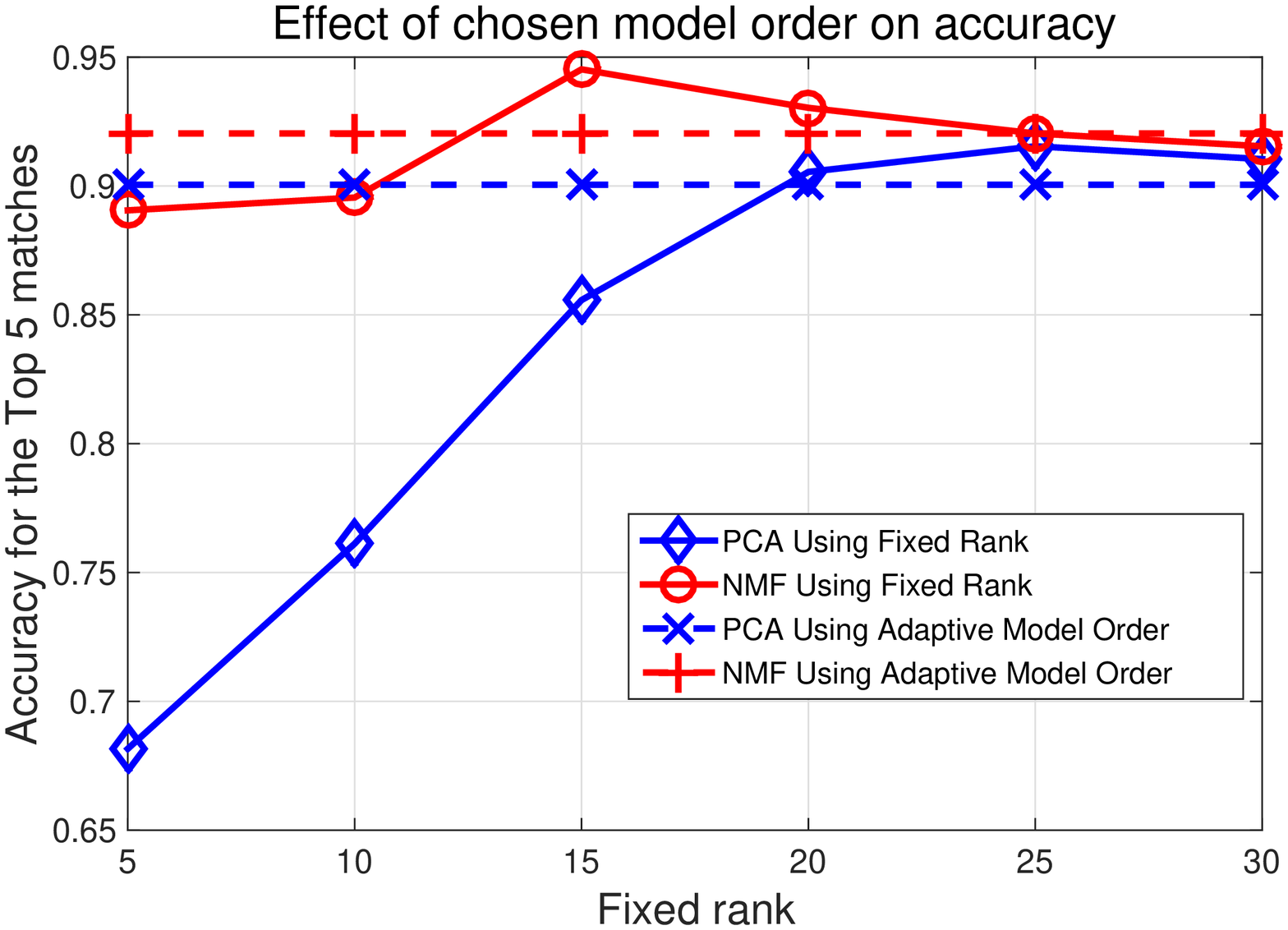}
\par\end{centering}
\caption{\label{fig:optimal_rank} Estimating the model order according to Section~\ref{sub:model_order}, incurs little or no penalty with 
respect to the best possible model order.}
\end{figure}

The effectiveness of the model order estimation from Section~\ref{sub:model_order} is examined in Figure~\ref{fig:optimal_rank}. Here, the rank of $\mathbf{H}$ and $\mathbf{L}$ for both query and server images was fixed at varying levels (x-axis). The retrieval accuracy was compared against 
that obtained using the estimated model order (horizontal lines).  Using the estimated model order
%under dimensionality reduction via PCA or NMF, when the reduced rank of $\mathbf{H}$ or $\mathbf{L}$ is computed adaptively, 
incurs little or no performance penalty relative to the fixed order schemes. 
%Note that selecting the optimal rank also facilitates the decision of what is the best possible dimensionality reduction unique for each image. 
Evidently, the method of Section~\ref{sub:model_order} provides a reasonable estimate of the model order, and consequently, the amount of query 
information sent to the server. The average model order, i.e., the average number of descriptors,  for the
ZuBuD images was 24.5980.
% \textbf{[Jacob: It is not clear what similarity metric has been used in these experiments. Furthermore, I am not sure how the accuracy of the fixed rank schemes is computed, in the context of the information displayed on the x-axis. I remember we discussed about this. However, that information has not been included here.]}

\begin{figure}[tb]
\begin{centering}
\includegraphics[width=8cm]{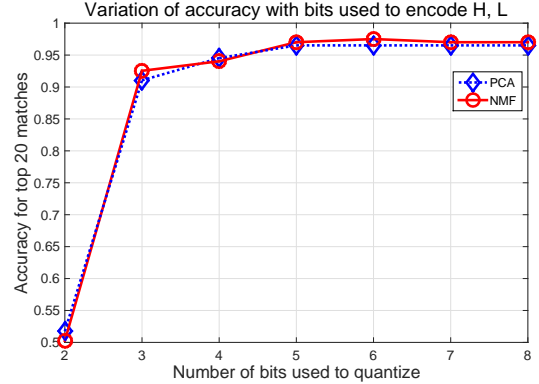}
\par\end{centering}
\caption{\label{fig:Quantization}Matching accuracy versus quantization rate (top 20 matches).}
\end{figure}

Next, we examine the impact of quantizing the elements of $\mathbf{H}$ and $\mathbf{L}$. 
Quantization limits the amount of query information uploaded to the server and reduces the memory required to store the 
image descriptors at the server. In Figure~\ref{fig:Quantization}, we examine the matching accuracy for different levels of fixed-rate quantization.
Using more than 5 bits to quantize the entries in $\mathbf{H}_q$, $\mathbf{L}_q$, $\mathcal{H}$ and $\mathcal{L}$ does not lead to further gains in 
matching accuracy. Thus, we employ 5-bit quantization to encode $\mathbf{H}_q$, $\mathbf{L}_q$, $\mathcal{H}$ and $\mathcal{L}$ in subsequent experiments. At this quantization level, the average size of the payload (per image) sent from the client to the server was 3.84 kilobytes 
for $\mathbf{H}_q$ and $\mathbf{L}_q$ combined.  In comparison, the method of~\cite{li2012quantized} requires only 2.5 kilobytes per image, but
incurs a server-based complexity quadratic in the number of SIFT features, as described in Section~\ref{sec:complexity}.
%\textbf{[Jacob: It is not clear here again what similarity metric has been used in these experiments. Furthermore, it is also unclear whether the information at the server has been quantized too, or this only holds for the query data. We need to clarify these missing aspects.]}

\begin{figure}[tb]
\begin{centering}
\includegraphics[width=8cm]{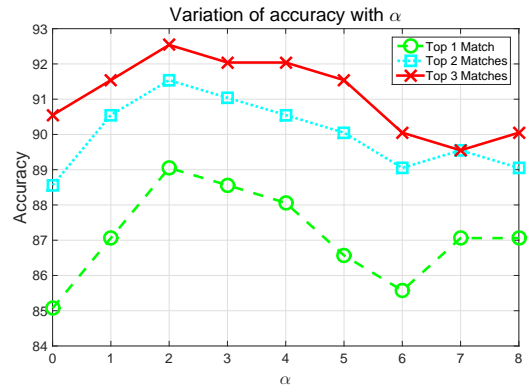}
\par\end{centering}
\caption{\label{fig:alpha}Varying $\alpha$ changes the relative weights of the primary (NMF) and secondary (PCA) hypotheses, altering overall
performance.}
\end{figure}

\begin{figure}[tb]
\begin{centering}
\includegraphics[width=8cm]{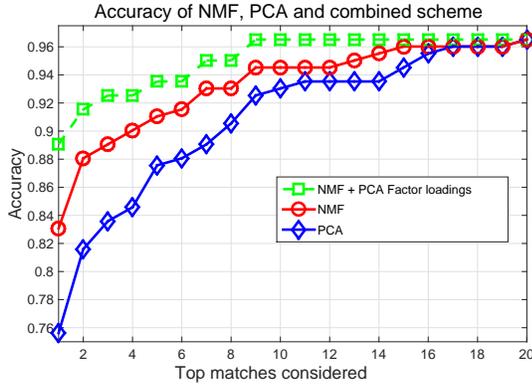}
\par\end{centering}
\caption{\label{fig:combined}The combined scheme gives more accurate retrieval than the separate PCA-based or sparse NMF-based approaches.}
\end{figure}

Using NMF and PCA individually leads to matching accuracies of 83\% and 75\% respectively, for the top 1 match.  
%Here, correlation between columns is used for comparing $\mathbf{H}_q$ with $\mathcal{H}$, angle between subspaces is used when comparing $\mathbf{L}_q$ with $\mathcal{L}$ and number of quantization bits set to 5. 
Now, we choose $\eta = 20$ and set NMF as the primary hypothesis, and PCA as the secondary hypothesis. The accuracy of the combined scheme is plotted in Figure~\ref{fig:alpha} for the top 1, 2, and 3 matches, as a function of the parameter $\alpha$. Recall that $\alpha$ serves to refine the primary hypothesis using the secondary hypothesis in Algorithm~\ref{alg:Combining-classifiers}. We find that $\alpha = 2$ leads to the best performance, though retrieval accuracy
does not decrease monotonically for $\alpha>2$. 
%This is because there is no clear way to weigh the contributions of primary and secondary hypothesis and it is highly dependent on the data and needs to be %obtained using empirical means.
% \textbf{[Jacob: The similarity metric used in these experiments is again not specified. We can be more descriptive when presenting these results.]}
Finally, with this combination of $\alpha=2$, 5-bit quantization, and the above similarity metrics, we examine the matching accuracy of all feature
clustering techniques for the top-1 to top-20 matches in Figure~\ref{fig:combined}. The combined scheme outperforms the two individual approaches, 
leading to a top 1, top 2 and top 3 matching accuracy of 89.05\%,  91.54\% and  92.54\% respectively. %\textbf{[Jacob: The similarity metric used in these experiments is again not specified. We can be more descriptive when presenting these results.]}

\section{Conclusions}
\label{sec:conc}
% !TEX root = ../MAIN.tex

PCA and sparse NMF were explored for feature clustering, i.e., reduction of the number of scale-invariant features extracted for content-based image retrieval. A measure of information content, previously used in econometrics, was employed to estimate the number of descriptors to be sent by the client device. For a database of urban images, 
%PCA and NMF-based schemes gave their best performance under different similarity metrics. 
combining the PCA and NMF approaches provides a top-3 matching accuracy of 92.54\%, which is competitive with previous work, but incurs significantly lower 
matching complexity due to feature clustering. Compared to pairwise matching based on thousands of native SIFT features per image, 
our method needs only about 25 PCA and NMF factor loadings per image. In ongoing work, we are extending this approach 
to other feature spaces (e.g., BRISK, FREAK, etc.), studying new theoretically motivated ways to estimate the model order, and examining new applications based on image classification.

\bibliographystyle{ieeetr}
\bibliography{refs2}

\end{document}